\documentclass{article}
\pdfoutput=1
\usepackage{ijcai19}
\usepackage{multirow}
\usepackage{bbm}

\usepackage{times}
\usepackage{algorithm}
\usepackage{algorithmic}
\usepackage{caption}
\usepackage{soul}
\usepackage{url}
\usepackage{graphicx}
\usepackage{amsmath}
\usepackage{booktabs}
\usepackage{algorithm}
\usepackage{algorithmic}
\usepackage[utf8]{inputenc}  
\usepackage{array} % Improves `tabular` and `array` environments
\usepackage{pict2e} % Allows \linethickness{...} in diagonal lines
\usepackage{slashbox} % Defines \backslashbox{..}{..}
\usepackage{natbib}
\usepackage{color, colortbl}
\usepackage[colorinlistoftodos]{todonotes}
\usepackage[T1]{fontenc}   
\usepackage{url}
\usepackage{booktabs}       % professional-quality tables
\usepackage{amsfonts}       % blackboard math symbols
\usepackage{nicefrac}       % compact symbols for 1/2, etc.
\usepackage{microtype}      % microtypography
\usepackage{diagbox} %backslashbox uses it for a diagonally split table cell
\usepackage{dsfont}
\usepackage{bm} % math bold. Example of use: %$\bm{BWT^{+}}$
\usepackage[colorlinks=true, allcolors=blue]{hyperref}
\usepackage{amsmath}
\usepackage{subcaption}
\usepackage{stmaryrd}
\usepackage{breakcites}
\usepackage{pdfpages}
\usepackage{rotating}
\usepackage{dirtree} %tree drawing

\title{%Neural-Symbolic reasoning for Explainable AI: \\ Toward a Reasoning Model}
Towards Explainable Neural-Symbolic Visual Reasoning} %image captioning}

\author{
Adrien Bennetot$^{1,2,3}$
\and
Jean-Luc Laurent$^1$\and
Raja Chatila$^2$\And
Natalia Díaz-Rodríguez$^3$
\affiliations
$^1$ Segula Technologies, Parc d'activité de Pissaloup - Trappes, France\\
$^2$ Institut des Systèmes Intelligents et de Robotique, Sorbonne Universite, France\\ %Campus Pierre et Marie CURIE, 
$^3$ U2IS \& INRIA FLOWERS \url{https://flowers.inria.fr} Team, ENSTA Paris, Palaiseau, France\\
\emails
\{adrien.bennetot, jeanluc.laurent\}@segula.fr,
raja.chatila@sorbonne-universite.fr,
natalia.diaz@ensta-paris.fr
}

\begin{document}

\maketitle

\begin{abstract}
Many high-performance models suffer from a lack of interpretability. There has been an increasing influx of work on explainable artificial intelligence (XAI) in order to disentangle what is meant and expected by XAI. Nevertheless, there is no general consensus on how to produce and judge explanations. In this paper, we discuss why techniques integrating connectionist and symbolic paradigms are the most efficient solutions to produce explanations for non-technical users and we propose a reasoning model, based on definitions by \citet{Doran17}, to explain a neural network's decision. We use this explanation in order to correct bias in the network's decision rationale. We accompany this model with an example of its potential use, based on the image captioning method in \citet{Burns18}.
\end{abstract}

\section{Existing perspectives in Explainable AI}

The last years have been characterized by an upsurge of opaque decision systems, such as Deep Neural Networks (DNN). Although they have great generalization and prediction skills, their functioning does not allow detailed explanations of their behaviour to be obtained. As opaque machine learning models are increasingly being employed to make important predictions in critical environments, the danger is to create and use decisions that are not justifiable or legitimate.

While it is not systematically necessary to obtain from the system an intelligible explanation, for example when the model has already been extensively studied and evaluated, the demand for interpretability is increasing from the various stakeholders in Artificial Intelligence. There is a general agreement on the importance of providing interpretability in machine learning models but desiderata differ according to the needs of each faction. Some common cases are a need for ethics~\citep{Goodman17}, for safety when using AI in high-risk environments ~\citep{Caruana15GAMS} or a need to allow the final user to trust the system~\citep{Zhu18}. The case we are interested in is that of developers trying to debug their models in order to make them more efficient, reliable and robust. It is customary to think that by focusing solely on performance, the systems will be increasingly opaque. This is true in the sense that there is a trade-off between the performance of a model and its transparency \citep{Dosilovic18}. However, an improvement in the understanding of a system can lead to the correction of its deficiencies. 

Since the definition of interpretability is subject to debate in the scientific community, we will use here the one proposed in \citet{Biran17}: 

\textit{Interpretability} is the \textit{degree to which an observer can understand the cause of a decision.} 
%\cnat{NeSy community normally speaks about NeSy learning and reasoning, here you need ot be specific with what you mean. NeSy XAI? NeSy learning and reasoning? i believe can be all of these, but avoid thorough NeSy as it is incomplete not saying the noun after}

Broad consensus exists on the importance of interpretability for AI models. However, there is no collective agreement on how to evaluate interpretation techniques. A division is regularly done between methods explaining the process of the model, called \textit{transparency} methods, versus \textit{post-hoc} methods. According to the Oxford dictionary, \textit{Post-hoc} reasoning is "occurring or done after the event, especially with reference to the fallacious assumption that the occurrence 
%\nat{of what? indicate. Inference occurrence?}\adr{it is a definition directly quoted from the dictionnary so no need to precise more} 
in question has a logical relationship with the event it follows". This is in contrast to the search for \textit{transparency}, which consists of a direct clarification of the model. 

\citet{Lipton18} call \textit{transparency} the opposition of black box-ness, the search for a direct understanding of the mechanism by which a model works. The contrast with \textit{post-hoc} methods, all those that do not clarify the model, is not exclusive, as a post-hoc method could also be considered as transparent if it provides an intuitive explanation of the model parameters. 
%Same definition as the latter 
%\nat{of []? indicate to make it clear}
%is provided by \citet{Lepri17}, adding the statement that post-hoc interpretations do not elucidate the exact process by which models work. 
The latter definition is also assumed by \citet{Montavon18}, and states, as a further remark, that the goal of post-hoc methods is to understand what the system predicts given a trained model. 
In \citet{Dosilovic18}, however, the authors add two sub-approaches to transparent methods: pure transparent ones, i.e., those that use model families considered as transparent such as linear models or decision trees, and hybrid ones that combine transparent model families with black box methods. This work also explains that human thinking is not transparent to us and, as human beings, we are used to justify our own decision thanks to a post-hoc mechanism, without fully knowing our decision process. 

Slightly different definitions behind these two terms are set in \citet{Preece18Stakeholders}, as they only place methods that do not derive from an internal state of the model (such as e.g., feature visualization~\citep{Olah18}), in the post-hoc category. This means that visualizing activations of different layers of a network is not purely post-hoc, but transparency-based. 
 
A distinction with more hindsight is made in \citet{Adadi18} by separating methods that take into consideration both the process and the outcome of the model on the one hand, and those that focus only on the outcome, on the other hand. They use the term of \textit{model-based} methods, which aims at understanding how the prediction process works as a transparency method, versus post-hoc methods. An explanation is considered as post-hoc if it is the produce of a separate method explaining the prediction produced by the black box while ignoring the decision process. 

Confronting transparency with post-hoc generates most of the time an opposition between symbolic and connectionist methods. On the one hand, symbolic methods are popularly considered less efficient, while they offer greater explainability. On the other hand, connectionist methods are more precise but opaque.

\section{The needs for neural-symbolic interpretability}

It has been proven that using a background knowledge within a DNN can bring robustness to the learning system \citep{Donadello17,donadello2018semantic,dAvilaGarcez19NeSy}. The use of a Knowledge Base (KB) to learn and reason with symbolic representation has the advantage of promoting the production of explanations while making a prediction \citep{Donadello19}. Neural-symbolic computation \citet{manhaeve2018deepproblog} is a promising path in order to move forward XAI

In his paper, \citet{Miller19} highlighted major findings that should be considered when creating an explainable AI model. First, explanations are better when constrictive, meaning that a prerequisite for a good explanation is that it does not only indicate why the model made a decision X, but also why it made decision X rather than decision Y. The ability to refer to established reasoning rules allows symbolic methods to fulfill this property. It is also explained in Miller's article that probabilities are not as important as causal links in order to provide a satisfying explanation. Considering that black box models tend to process data in a quantitative manner, it would be necessary to translate the probabilistic results into qualitative notions containing causal links. Again, the use of symbols could carry this property as the use of a knowledge base (KB) such as an ontology can allow data to be processed directly in a qualitative way. In addition, they state that explanations are selective, meaning that focusing solely on the main causes of a decision-making process is sufficient. It is known that there is a trade-off between interpretability and accuracy \citep{Gilpin18}, i.e., between the simplicity of the information given by the system on its internal functioning, and the exhaustiveness of this description. Considering that additional variables and equations must be introduced in order to test whether a correlation between two variables is genuine or spurious \citep{Herbert12}, being selective is less straight-forward for connectionist models than for symbolic ones. Finally, considering that a good explanation needs to influence the mental model of the user, i.e. the representation of the external reality using, among other things, symbols, it seems obvious %to us 
that the use of the symbolic learning paradigm is appropriate to produce an explanation. 

One of the goals of having interpretability in a model is to explain its reasoning by expressing it in a way that is understandable and readable by human beings, while highlighting the biases learned by the model, in order to validate or invalidate its decision rationale \citep{Guidotti19}. It is customary to think that by focusing solely on performance, the systems will be increasingly opaque. This is true in the sense that there is a trade-off between the performance of a model and its transparency \citep{Dosilovic18}. We consider that the advocacy for interpretability may lead to a generic performance improvement for 3 reasons: i) it will help ensure impartiality in decision-making, i.e. to highlight, and consequently, correct from bias in the training data-set, ii) interpretability facilitates the provision of robustness by highlighting potential adversarial perturbations that could change the prediction, and finally, iii) interpretability can act as an insurance that only meaningful variables infer the output, i.e., guaranteeing that an underlying truthful causality exists in the model reasoning. Combining the prediction capabilities of connectionist models with the transparency of symbolic ones will put aside the trade-off by increasing both the interpretability and the performance of AI models.

Therefore, neural-symbolic interpretability can provide convincing explanations while keeping or improving generic performance.  

\section{Neural-Symbolic computation for truly Explainable AI}

Truly explainable models should directly integrate reasoning, in order to not leave explanation generation to the human user. In the model proposed by \citet{Doran17}, the black box, i.e. the connectionist part, is giving the final output, while the KB is externally provided to the model. This allows the system to generate itself an explanation in natural language, thus linking the high level features identified by the model and the final output. It also highlights the logical path the model should have taken: since the KB is given by the user and (therefore we assume) cannot be incorrect, a reasoning error in the natural language explanation would signify a mistake in the black box between high level features and the final output. In addition, as stated in \citet{Doran17}, the inclusion of reasoning in the model eliminates the potential corruption of the explanation that could arise from using external sources to justify the actual model we want to make explainable. 

%TODO First hypotehsis experiment to do is: what provides a better quality explanation for the BB and for the user, a externally or internally created KB? this we are obviating but perhaps, despite inducing bias, the explanation is more satisfactory with external bias, or a compromise of the two to explain the missing links while preserving entities and relations that are from the actual BB only. #EXPERIMENT TO DO.

However, we can propose some adjustments in this architecture: the causal links given by the KB do not directly reflect the operations that took place in the black box, and it is therefore impossible to affirm that the model predicted this output for the reasons given in the natural language explanation. Since nothing connects the KB and the black box, therefore it is impossible to link the explanation and the predicted output. The objective of not leaving explanation generation to human analysts is fulfilled, as the model formulates a line of reasoning, but the explanation given is not correct (or does not have accurate provenance), as it only explains what the black box should have learned, and not what it actually learned.

A possible adaptation would be to not use the output of the black box in the reasoner and solely use the high level features detected by the model so that the natural language explanation would match the reasoning that led to this result. This would mean truncating the potential of the black box. It is possible to link the reasoner and the black box by considering that the output is no longer the final result, but rather high level features. The model would then produce an explanation on what the system should conclude when seeing those features but not why it detected those features.

A last option to achieve an explanation of the model decision would be to directly populate the KB from the data. This would allow to provide an explanation in natural language directly from the black box, emphasizing in the meantime the model's reasoning errors and highlighting possible bias in the dataset or model. This is the option we propose pursuing as we believe it provides the most faithful explanation of how the model actually works.

We summarize the different scenarios in Table \ref{fig:scenarios}.

\begin{table*}[]
\begin{tabular}{ll|l|l|l}
\cline{3-4}
 &  & \multicolumn{2}{c|}{%Origin of the Knowledge Base  Provenance is a term used in KG literature
 Knowledge Base Provenance} &  \\ \cline{3-4}
 &  & \multicolumn{1}{c|}{%Given by the user
 External} & \multicolumn{1}{c|}{Black box} &  \\ \cline{1-4}
\multicolumn{1}{|c|}{\multirow{2}{*}{\begin{tabular}[c]{@{}c@{}}\\%Origin of the model's final output
Model's final output origin
\end{tabular}}} & \multicolumn{1}{c|}{\begin{tabular}[c]{@{}c@{}}Reasoner\end{tabular}} & \begin{tabular}[c]{@{}l@{}} - No explanation about the black box  \\ - Does not highlight reasoning mistakes\end{tabular} & \multirow{2}{*}{\begin{tabular}[c]{@{}l@{}}\textbf{+ Explanation about the black box}  \\\textbf{+  Highlights reasoning mistakes %(e.g. bias)
} \\
%+ Detect dataset bias faithfully 
\end{tabular}} & \multirow{2}{*}{} \\ \cline{2-3}
\multicolumn{1}{|c|}{} & \multicolumn{1}{c|}{Black box} & \begin{tabular}[c]{@{}l@{}}  - No explanation of the black box \\ + Highlight reasoning mistakes \\ \end{tabular} &  &  \\ \cline{1-4}
\end{tabular}
\caption{Scenarios for Neural-Symbolic Reasoning, depending on the origin of the KB and the origin of the final output. The cell with text in bold is our contributed proposed model to achieve faithful neural-symbolic visual reasoning.}
\label{fig:scenarios}
\end{table*}

We derive two prerequisites that are necessary to create a truly reasoning AI: i) The KB must inherently emerge from the black box model in order to conceptually (symbolically) reflect what the model learned. ii) The symbolic part must constraint %in the future, also: and close the loop with 
the connectionist part to improve the prediction performance of the model.  
% TODO: N: This is what is not end-to-end differentiable yet, and we need to do, cannot be hardcoded. See Angela Fan template generation of stories, how they do it end to end differentiable (hierarchical story generation)

\begin{figure}[h!]
\centering
\includegraphics[scale=0.25]{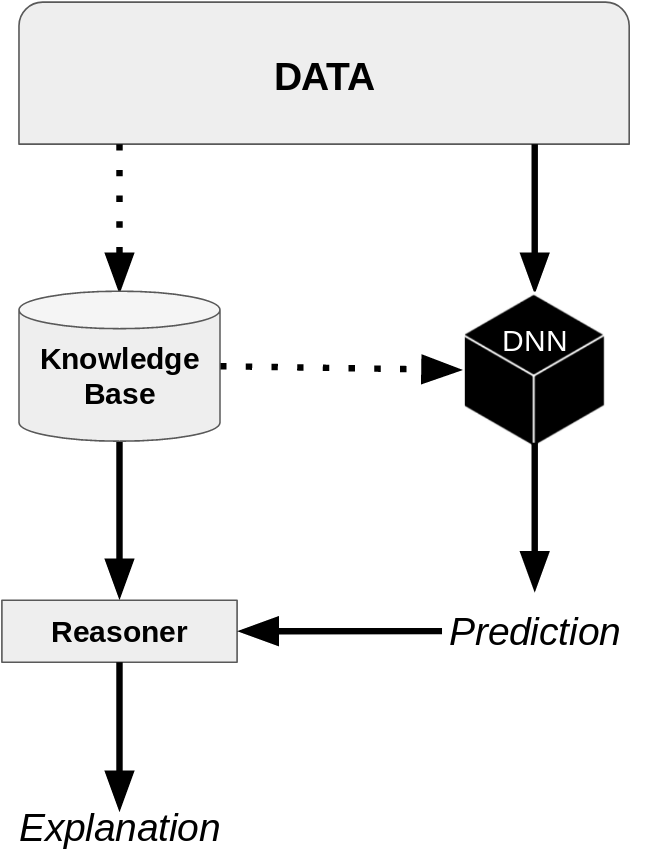} 
\caption{Proposed neural-symbolic explainable model extended from \citet{Doran17}: the black box model provides, along with its output, an explanation of its reasoning to highlight bias and improve performance. Our contribution with respect to \citet{Doran17} is the way we populate the KB directly from the data and the way we constraint the DNN thanks to the KB. It can be seen with the dashed lines}
\label{fig:personal_production}
\end{figure}

We propose an adaptation of the architecture in Fig. \ref{fig:personal_production}. Instead of externally providing a KB to complement the model, we propose to i) directly extract symbolic rules from a first black box and ii) reflect those rules in a second black box by constraining learning according to perceived properties, e.g., by modifying initialization protocols, loss functions or hyperparameters. Therefore, the model's ultimate output would come from the reasoner but would be directly influenced by both the black box and the KB, i.e., it would not truncate the black box potency but reveal, as expected by an explanation, the biases learned by the model and lead to performance improvement while explaining in natural language its prediction. 

As it seems intuitive that the presence of a KB is useful to provide an explanation, how to use it to influence a network raises some questions. The role of the reasoner is to answer to queries based on the \textit{prediction} and whose answers come from the symbolic rules stored in the KB. 

\section{Towards a model for XAI through neural-symbolic computation }

One barrier to transparency is a "mismatch between the mathematical optimization using high-dimensionality characteristics of machine learning and the demands of human-scale reasoning and styles of interpretation" \citep{Burrell16}. With the objective of reducing this gap, and inspired by the work of \citet{Burns18}, we hypothesize that the use of loss functions that have a concrete and more graspable perceptible meaning could make it easier to provide an explanation than a classic non intuitive cross-entropy. In \citet{Burns18}, authors introduce two new loss functions: the "Appearance Confusion Loss" and the "Confident Loss" in order to counter-balance gender bias during an image captioning process. The Appearance Confusion Loss is based on the fact that \textit{for an image devoid of gender information, the probability of predicting man or woman should be equal}; and the Confident Loss exists to encourage the model to predict gender words correctly when gender evidence is present. 

In order to test our model, we address the problem of reducing the bias in image captioning explained in \citet{Burns18}. In a succinct manner, it consists in training a neural network using images \textit{I}, image captions S, and image segmentation annotation masks \textit{M}, with a Neural Image Caption network \citep{Bengio14} as a base. We chose this problem because it represents a case where the creation of a KB from the data is possible, and because captioning is vulnerable to bias. 

We propose creating the reasoning-facilitating KB by performing word-embedding on the black box model labels in order to determine which words are particularly exposed to a risk of errors due to learning priors or biased data collection. Interchangeable words are more likely to be victims of overuse of context \citep{Zhao17}. 
 
To show a trivial example, if seniors are represented with park benches more often than young adults, the model may tend to misuse the context by predicting a senior each time a bench is detected, without further caring about the person on the picture. Our explainable model proposes extracting a list noted as $B_{word}$ and used in a similar syntax layout than a chosen word, such as $B_{person} = [\textit{man, teenager, boy, senior}]$. This means that \textit{man}, \textit{teenager}, \textit{boy} and \textit{senior} are often used in the same context as the word \textit{person}. The extracted list will constitute a set of words base for which the model will have to hesitate when trying to predict one word rather than another. We call \textit{person} a class while \textit{man}, \textit{teenager}, \textit{boy} and \textit{senior} are ontological sub-classes. We want to force the system to hesitate on which ontological sub-class it should predict when a class is present on an image, in order to not make any mistakes, but to be confident enough to nevertheless predict an ontological sub-class rather than a class, in order to be as specific as possible.  

We propose to generalize the gender specific loss functions from \citet{Burns18} in an attempt to force the system to be confused (or confident, resp.) when predicting any \textit{bias-prone} word, i.e, any ontological sub-class from the set % from 
$B_{word}$ belonging to the \textit{word} class.

Just like \citet{Burns18}, we want to force our model to be confused when making predictions if the input image does not contain appropriate evidence for the prediction to be made. We use masked images $I'$, where the information relevant to making a decision, such as the interior of a segmentation mask for a human in the image (if we are trying to determine whether or not the human classified is a senior or not), is removed. To ensure equiprobability among the different words in $B_{word}$ when there is no appropriate information for the system to predict a specific word, but rather its generic category, we generalize the confusion function from \citet{Burns18} to the general case where gender is not the only source of potential bias. We note confusion function $C$ which operates over the predicted distribution of words $p(\tilde{w}_t$):

\begin{equation}
   C(\tilde{w}_t,I') = \sum_{b \in B_{word}} (p(\tilde{w}_t = b | w_{0:t-1}, I') - \frac{1}{J})^2
\end{equation}

where $J$ is the length of $B_{word}$. As we try to minimize $C(\tilde{w}_t,I')$, we have a sum of squares that tends toward zero. Given that if a sum of squares is zero, each term must be zero, each probability tends to be equal. As in \citet{Burns18}, we define the confusion loss $\mathcal{L}^{Confusion}$ as: 

\begin{equation}
  \mathcal{L}^{Confusion} = \frac{1}{N}\sum_{n=0}^{N}\sum_{t=0}^{T}\mathbbm{1} (w_t \in B_{word})C(\tilde{w}_t,I'),
\end{equation}

with $\mathbbm{1}$ an indicator variable that denotes whether or not $w_t$ is a bias-prone word, $N$ the batch size, and $T$ the number of words in the given sentence. As we want the model to be confident about its prediction when there is an appropriate information on the image, this time we use complete (i.e., non masked) images $I$ as input instead of masked ones $I'$. With $j$ the index of word $b$ in list $B_{word}$, we have the confidence function $F^{j}$.  

\begin{equation}
   F^{j}(\tilde{w}_t, I) = \frac{\sum_{b \in B_{word}\backslash b_j} p(\tilde{w}_t = b|w_{0:t-1}, I)}{p(\tilde{w}_t = b_j|w_{0:t-1}, I) + \epsilon}
\end{equation}

As in \citet{Burns18}, we add an $\epsilon$ for numerical stability. $F^{j}$ will tend towards zero if $p(\tilde{w}_t = b_j$) dominates the sum of the predicted distribution of every other \textit{bias-prone} word.   

We use $F^{j}$ to define the confident loss $\mathcal{L}^{Confidence}$:

\begin{equation}
\mathcal{L}^{Confidence} = \frac{1}{N}\sum_{n=0}^{N}\sum_{t=0}^{T}\sum_{j=1}^{J}(\mathbbm{1}(w_t = b_j)F^{j}(\tilde{w}_t, I))
\end{equation}

By adding a standard cross-entropy loss $\mathcal{L}^{CE}$ to non-bias-prone words, we obtain a model able to use context priors when there is no interchangeable word for the predicted one and to be confused/confident when the question arises thanks to the loss $\mathcal{L}$:

\begin{equation}
\mathcal{L} = \alpha\mathcal{L}^{CE} + \beta\mathcal{L}^{Confidence} + \mu\mathcal{L}^{Confusion}
\end{equation}

with $\alpha$, $\beta$ and $\mu$ hyper-parameters.

\begin{figure}[h!]
\centering
\includegraphics[scale=0.35]{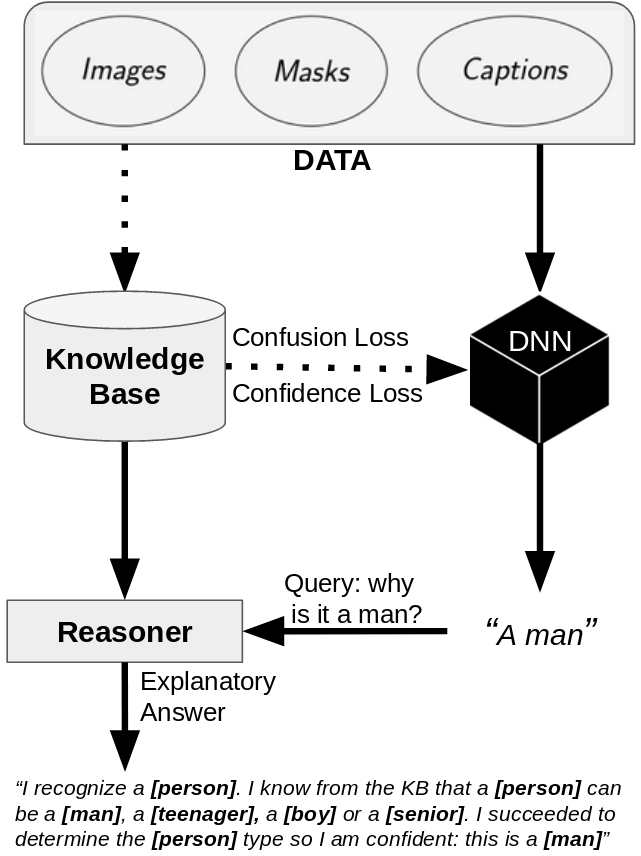}
\caption{Example of use of our neural-symbolic explainable model. The black box gives the caption output while the reasoner gives captioning explanation}
\label{fig:personal_production2}
\end{figure}

The reasoner can provide a state-based explanation of the learning of the neural network, depending on the output result: i.e., it can naturally provide a \textit{confident explanation state} if it succeeds to predict an ontological sub-class or a \textit{confused explanation state} if it predicted a class, as shown in Fig. \ref{fig:personal_production2}. 

This model, when applied to image captioning or object recognition tasks, has several advantages: i) it detects the provenance of bias in a black box model such as a neural network, ii) gives an unbiased prediction for which the context has not been overused, and iii) gives an explanation in natural language on the neural network's functioning; particularly, on its loss-based optimization procedure.

\section{Conclusion}
%TODO: We build upon Derek to further characterize what a NeSy XAI model output could produce.
Models combining connectionism and symbolism are not widely represented in the state of the art of XAI. These paradigms are rarely combined when providing explanations. The use of a symbolic basis with a neural network can provide explanations close to the functioning of human reasoning while %keeping at the same time high performance
maintaining the state-of-the-art performance at the same time. We build upon \cite{Doran17} and extend \cite{Burns18} to further characterize what a neural-symbolic explainable model could output. We propose a model endowed with a non-external KB, i.e., directly built on the learning %training 
%N: in future we could also use test data to keep continual learning, not sure if we have space to properly say it here, or save it to do it ourselves later
data of a neural network, that allows to influence its learning and to correct bias thoroughly, while giving a fair explanation from its predictions. As the user or expert external knowledge does not interfere the predictions in the explanation process, it constitutes a truly explainable model that is faithful to communicate the reasoning behind its output decisions. Future work will evaluate and challenge %TODO extend the model to handle relationships as well as entities, but also relationships, as well as 
the model in realistic datasets.

%%ToDO: "state-based" "reasoner"

\bibliographystyle{named}
\bibliography{references}

\begin{thebibliography}{}

\bibitem[\protect\citeauthoryear{{Adadi} and {Berrada}}{2018}]{Adadi18}
Amina {Adadi} and Mohammed {Berrada}.
\newblock Peeking inside the black-box: A survey on explainable artificial
  intelligence (xai).
\newblock {\em IEEE Access}, 6:52138--52160, 2018.

\bibitem[\protect\citeauthoryear{Biran and Cotton}{2017}]{Biran17}
Or~Biran and Courtenay Cotton.
\newblock Explanation and justification in machine learning: A survey.
\newblock In {\em IJCAI-17 workshop on explainable AI (XAI)}, volume~8, page~1,
  2017.

\bibitem[\protect\citeauthoryear{{Burns} \bgroup \em et al.\egroup
  }{2018}]{Burns18}
Kaylee {Burns}, Lisa~Anne {Hendricks}, Kate {Saenko}, Trevor {Darrell}, and
  Anna {Rohrbach}.
\newblock {Women also Snowboard: Overcoming Bias in Captioning Models}.
\newblock {\em arXiv e-prints}, page arXiv:1803.09797, Mar 2018.

\bibitem[\protect\citeauthoryear{Burrell}{2016}]{Burrell16}
Jenna Burrell.
\newblock How the machine ‘thinks’: Understanding opacity in machine
  learning algorithms.
\newblock {\em Big Data \& Society}, 3(1):2053951715622512, 2016.

\bibitem[\protect\citeauthoryear{Caruana \bgroup \em et al.\egroup
  }{2015}]{Caruana15GAMS}
Rich Caruana, Yin Lou, Johannes Gehrke, Paul Koch, Marc Sturm, and Noemie
  Elhadad.
\newblock Intelligible models for healthcare: Predicting pneumonia risk and
  hospital 30-day readmission.
\newblock In {\em Proceedings of the 21th ACM SIGKDD International Conference
  on Knowledge Discovery and Data Mining}, KDD '15, pages 1721--1730, New York,
  NY, USA, 2015. ACM.

\bibitem[\protect\citeauthoryear{d'Avila Garcez \bgroup \em et al.\egroup
  }{2019}]{dAvilaGarcez19NeSy}
Artur~S. d'Avila Garcez, Marco Gori, Lu{\'{\i}}s~C. Lamb, Luciano Serafini,
  Michael Spranger, and Son~N. Tran.
\newblock Neural-symbolic computing: An effective methodology for principled
  integration of machine learning and reasoning.
\newblock {\em CoRR}, abs/1905.06088, 2019.

\bibitem[\protect\citeauthoryear{Donadello \bgroup \em et al.\egroup
  }{2017}]{Donadello17}
Ivan Donadello, Luciano Serafini, and Artur~D'Avila Garcez.
\newblock Logic tensor networks for semantic image interpretation.
\newblock {\em Proceedings of the Twenty-Sixth International Joint Conference
  on Artificial Intelligence, IJCAI}, pages 1596--1602, 2017.

\bibitem[\protect\citeauthoryear{Donadello \bgroup \em et al.\egroup
  }{2019}]{Donadello19}
Ivan Donadello, Mauro Dragoni, and Claudio Eccher.
\newblock Persuasive explanation of reasoning inferences on dietary data.
\newblock In {\em First Workshop on Semantic Explainability @ ISWC 2019}, 2019.

\bibitem[\protect\citeauthoryear{Donadello}{2018}]{donadello2018semantic}
Ivan Donadello.
\newblock {\em Semantic image interpretation-integration of numerical data and
  logical knowledge for cognitive vision}.
\newblock PhD thesis, University of Trento, 2018.

\bibitem[\protect\citeauthoryear{{Doran} \bgroup \em et al.\egroup
  }{2017}]{Doran17}
Derek {Doran}, Sarah {Schulz}, and Tarek~R. {Besold}.
\newblock {What Does Explainable AI Really Mean? A New Conceptualization of
  Perspectives}.
\newblock {\em arXiv e-prints}, page arXiv:1710.00794, Oct 2017.

\bibitem[\protect\citeauthoryear{{Došilović} \bgroup \em et al.\egroup
  }{2018}]{Dosilovic18}
Filip~Karlo {Došilović}, Mario {Brčić}, and Nikica {Hlupić}.
\newblock Explainable artificial intelligence: A survey.
\newblock In {\em 2018 41st International Convention on Information and
  Communication Technology, Electronics and Microelectronics (MIPRO)}, pages
  0210--0215, May 2018.

\bibitem[\protect\citeauthoryear{{Gilpin} \bgroup \em et al.\egroup
  }{2018}]{Gilpin18}
Leilani~H. {Gilpin}, David {Bau}, Ben~Z. {Yuan}, Ayesha {Bajwa}, Michael
  {Specter}, and Lalana {Kagal}.
\newblock {Explaining Explanations: An Overview of Interpretability of Machine
  Learning}.
\newblock {\em arXiv e-prints}, page arXiv:1806.00069, May 2018.

\bibitem[\protect\citeauthoryear{Goodman and Flaxman}{2017}]{Goodman17}
Bryce Goodman and Seth Flaxman.
\newblock European union regulations on algorithmic decision-making and a
  "right to explanation".
\newblock {\em AI Magazine}, 38:50--57, 2017.

\bibitem[\protect\citeauthoryear{Guidotti \bgroup \em et al.\egroup
  }{2018}]{Guidotti19}
Riccardo Guidotti, Anna Monreale, Salvatore Ruggieri, Franco Turini, Fosca
  Giannotti, and Dino Pedreschi.
\newblock A survey of methods for explaining black box models.
\newblock {\em ACM Comput. Surv.}, 51(5):93:1--93:42, August 2018.

\bibitem[\protect\citeauthoryear{Lipton}{2018}]{Lipton18}
Zachary~C. Lipton.
\newblock The mythos of model interpretability.
\newblock {\em Queue}, 16(3):30:31--30:57, June 2018.

\bibitem[\protect\citeauthoryear{Manhaeve \bgroup \em et al.\egroup
  }{2018}]{manhaeve2018deepproblog}
Robin Manhaeve, Sebastijan Dumancic, Angelika Kimmig, Thomas Demeester, and Luc
  De~Raedt.
\newblock Deepproblog: Neural probabilistic logic programming.
\newblock In S.~Bengio, H.~Wallach, H.~Larochelle, K.~Grauman, N.~Cesa-Bianchi,
  and R.~Garnett, editors, {\em Advances in Neural Information Processing
  Systems 31}, pages 3749--3759. Curran Associates, Inc., 2018.

\bibitem[\protect\citeauthoryear{Miller}{2019}]{Miller19}
Tim Miller.
\newblock Explanation in artificial intelligence: Insights from the social
  sciences.
\newblock {\em Artif. Intell.}, 267:1--38, 2019.

\bibitem[\protect\citeauthoryear{Montavon \bgroup \em et al.\egroup
  }{2018}]{Montavon18}
Grégoire Montavon, Wojciech Samek, and Klaus-Robert Müller.
\newblock Methods for interpreting and understanding deep neural networks.
\newblock {\em Digital Signal Processing}, 73:1--15, 02 2018.

\bibitem[\protect\citeauthoryear{Olah \bgroup \em et al.\egroup
  }{2018}]{Olah18}
Christopher Olah, Arvind Satyanarayan, Ian Johnson, Shan Carter, Ludwig
  Schubert, Katherine Ye, and Alexander Mordvintsev.
\newblock The building blocks of interpretability.
\newblock {\em Distill}, 2018.

\bibitem[\protect\citeauthoryear{{Preece} \bgroup \em et al.\egroup
  }{2018}]{Preece18Stakeholders}
Alun {Preece}, Dan {Harborne}, Dave {Braines}, Richard {Tomsett}, and Supriyo
  {Chakraborty}.
\newblock {Stakeholders in Explainable AI}.
\newblock {\em arXiv e-prints}, page arXiv:1810.00184, Sep 2018.

\bibitem[\protect\citeauthoryear{Simon}{1954}]{Herbert12}
Herbert~A. Simon.
\newblock Spurious correlation: A causal interpretation*.
\newblock {\em Journal of the American Statistical Association},
  49(267):467--479, 1954.

\bibitem[\protect\citeauthoryear{{Vinyals} \bgroup \em et al.\egroup
  }{2014}]{Bengio14}
Oriol {Vinyals}, Alexander {Toshev}, Samy {Bengio}, and Dumitru {Erhan}.
\newblock {Show and Tell: A Neural Image Caption Generator}.
\newblock {\em arXiv e-prints}, page arXiv:1411.4555, Nov 2014.

\bibitem[\protect\citeauthoryear{{Zhao} \bgroup \em et al.\egroup
  }{2017}]{Zhao17}
Jieyu {Zhao}, Tianlu {Wang}, Mark {Yatskar}, Vicente {Ordonez}, and Kai-Wei
  {Chang}.
\newblock {Men Also Like Shopping: Reducing Gender Bias Amplification using
  Corpus-level Constraints}.
\newblock {\em arXiv e-prints}, page arXiv:1707.09457, Jul 2017.

\bibitem[\protect\citeauthoryear{Zhu \bgroup \em et al.\egroup }{2018}]{Zhu18}
Jichen Zhu, Antonios Liapis, Sebastian Risi, Rafael Bidarra, and
  Gregory~Michael Youngblood.
\newblock Explainable ai for designers: A human-centered perspective on
  mixed-initiative co-creation.
\newblock {\em 2018 IEEE Conference on Computational Intelligence and Games
  (CIG)}, pages 1--8, 2018.

\end{thebibliography}

\end{document}